\documentclass[10pt,twocolumn,letterpaper]{article}

\usepackage{iccv}
\usepackage{times}
\usepackage{epsfig}
\usepackage{graphicx}
\usepackage{amsmath}
\usepackage{amssymb}
\usepackage{multicol, multirow}

\renewcommand{\thefootnote}{\fnsymbol{footnote}}
\newcommand\blfootnote[1]{%
  \begingroup
  \renewcommand\thefootnote{}\footnote{#1}%
  \addtocounter{footnote}{-1}%
  \endgroup
}

\usepackage[breaklinks=true,bookmarks=false]{hyperref}

\iccvfinalcopy 

\def\iccvPaperID{****} 

\ificcvfinal\fi

\begin{document}

\title{Hybrid model for Single-Stage Multi-Person Pose Estimation}

\author{Jonghyun Kim$^\dagger$, Bosang Kim$^\dagger$, Hyotae Lee, Jungpyo Kim, Wonhyeok Im, \\ Lanying Jin, Dowoo Kwon, and Jungho Lee\\
LG Electronics\\
19, Yangjae-daero 11gil, Seocho-Gu, Seoul 06772, Republic of Korea\\
{\tt\small \{jonghyun0.kim, bosang1.kim, hyotae.lee, jungpyo.kim, wonhyeok.im,} \\
{\tt\small lanying.jin, dowoo.kwon, jungo.lee\}@lge.com}
}

\maketitle
\ificcvfinal\thispagestyle{empty}\fi

\begin{abstract}
   In general, human pose estimation methods are categorized into two approaches according to their architectures: regression (i.e., heatmap-free) and heatmap-based methods. The former one directly estimates precise coordinates of each keypoint using convolutional and fully-connected layers. Although this approach is able to detect overlapped and dense keypoints, unexpected results can be obtained by non-existent keypoints in a scene. On the other hand, the latter one is able to filter the non-existent ones out by utilizing predicted heatmaps for each keypoint. Nevertheless, it suffers from quantization error when obtaining the keypoint coordinates from its heatmaps. In addition, unlike the regression one, it is difficult to distinguish densely placed keypoints in an image. To this end, we propose a hybrid model for single-stage multi-person pose estimation, named HybridPose, which mutually overcomes each drawback of both approaches by maximizing their strengths. Furthermore, we introduce self-correlation loss to inject spatial dependencies between keypoint coordinates and their visibility. Therefore, HybridPose is capable of not only detecting densely placed keypoints, but also filtering the non-existent keypoints in an image. Experimental results demonstrate that proposed HybridPose exhibits the keypoints visibility without performance degradation in terms of the pose estimation accuracy. \blfootnote{$\dagger$ Both authors are equally contributed.}
   
\end{abstract}

\section{Introduction}
\label{sec:intro}

Multi-Person Pose Estimation (MPPE) aims to not only detect every person in a scene but also locate skeletal keypoints of each person. It has emerged as a popular research area in computer vision since the skeletal keypoints help understanding human activity \cite{duan2022revisiting, duan2022pyskl, chen2021channel}, human-object interaction \cite{zhang2022improved, zheng2020skeleton}, and so on. Typically, MPPE can be summarized into top-down, bottom-up, and single-stage approaches according to pose estimation procedures. The top-down approach first detects humans in a scene, and estimates keypoints of each detected human. Oppositely, the bottom-up method detects all keypoints in the scene at once, and groups them by individual persons. Finally, the single-stage approach simultaneously detects humans and their corresponding keypoints in a single step.

We reorganize MPPE into two categories in terms of its output to tackle ill-posed problems: \textit{i) Regression-based} \cite{carreira2016human, zhou2019objects, wei2020point, mcnally2022rethinkingkapao} and \textit{ii) Heatmap-based} \cite{jain2013learning, tompson2014joint, tompson2015efficient, wei2016convolutional, newell2016stacked, pishchulin2016deepcut, insafutdinov2016deepercut, chen2018cascadedCPN, sun2019deephrnet, cheng2020higherhrnet}. In the regression-based methods, coordinates of body keypoints are directly estimated as a network output. This property allows that these methods separately distinguish overlapped and dense keypoints of the same class. However, these methods yield non-existent keypoints in a scene since their output has a fixed dimension. 
Meanwhile, the heatmap-based methods first predict a set of probability maps for the each keypoint of interest, and then their locations can be obtained by finding the highest activation in each heatmap. Thereby, the non-existent keypoints can be filtered by predicted probability maps. However, undesirable effects still appear in the results by converting the heatmap to keypoint coordinates. When several keypoints of the same class are densely located, the heatmap-based methods struggle with distinguishing them since the distributions of their activation are overlapped. Moreover, the large-sized heatmaps are required to reduce quantization errors.


To solve the aforementioned issues, we propose a hybrid model for the MPPE, named HybridPose, which mutually compensates for each weakness adopting strengths of both regression and heatmap-based methods. To maximizing both strengths, HybridPose adopts a single-stage MPPE architecture to simultaneously predict human bounding boxes, keypoint coordinates, and further visibility maps. Specifically, HybridPose directly yields keypoint coordinates, and these are identified along with each person. After that, the visibility maps filter non-existent keypoints out by representing their existence as probability distributions. In addition, we propose self-correlation loss to align these distributions with their corresponding keypoint coordinates. As the independent phase, HybridPose predicts the keypoint locations and their visibility respectively. To associate both them with each one, the self-correlation loss injects spatial relationships between keypoint coordinates and visibility maps. Consequently, HybridPose is capable of estimating overlapped keypoints of the same class separately while eliminating non-existent ones.

We provide extensive experiments to verify the effectiveness of HybridPose. Moreover, we evaluate the performance of the proposed network in terms of various evaluation metrics. Experimental results demonstrate that the proposed HybridPose disposes invisible keypoints effectively without significant degradation of performance. Compared with existing methods, our contributions can be summarized as follows:

\begin{itemize}
    \item We propose a hybrid model for the multi-person pose estimation, named HybridPose, that maximizes virtues of both regression and heatmap-based methods to overcome each drawback.
    \item We provide self-correlation loss to construct spatial dependencies between keypoint coordinates and visibility maps.
    \item HybridPose simultaneously estimates human bounding boxes, keypoint coordinates, and further visibility maps in a single-stage network.
    \item With the help of HybridPose, chronic problems of both regression and heatmap-based methods are mitigated.
\end{itemize}

\section{Related Work}
\label{sec:relate}
\textbf{Regression-based methods}
DeepPose \cite{toshev2014deeppose} first introduced regression of keypoint coordinates using iterative deep neural networks. In addition, Joao \etal \cite{carreira2016human} proposed a network using Iterative Error Feedback (IEF). The IEF can improve the pose accuracy by finding inaccurate output obtained from the previous estimation scheme and correcting it iteratively. CenterNet proposed by Zhou \etal \cite{zhou2019objects} adopted pose regression in the single-shot object detection framework. By considering the each keypoint as a offset from the central point on a detection box, CenterNet can estimate the keypoints directly without pre-processing such as the region proposal. Consequently, CenterNet can achieve the MPPE, since then most of the single-stage MPPE methods have adopted regression-based approach. However, since its image features are human center oriented, these features have limited information for predicting distant keypoints, due to the variation of the human's height and their pose. To resolve this limitation, \cite{wei2020point,mcnally2022rethinkingkapao} adopted displacement of the pre-defined pose anchors to estimate each keypoint. Although existing regression-based methods have been proposed to increase the pose accuracy, it is still remained to estimating keypoints visibility as a ill-pose problem. Therefore, these regression-based methods can obtain unexpected results by the non-existent keypoints in a scene.

\textbf{Heatmap-based methods}
In \cite{jain2013learning, tompson2014joint}, the notion of heatmaps appeared as a spatial model by representing them with probability distributions. To develop the spatial model, Tompson \etal \cite{tompson2015efficient} utilized a cascade neural network to obtain refined heatmaps from coarse ones. However, these methods are able to yield keypoints of an only single person. To estimate joints of multiple persons in a scene, DeepCut \cite{pishchulin2016deepcut} adopted two-stage architectures. This method first detected human bounding boxes in the scene. After that, heatmaps were estimated per detected boxes. Oppositely, DeeperCut \cite{insafutdinov2016deepercut} first estimated all keypoints in the scene, and aligned them to each person. Although these methods achieved multi-person keypoint detection, the resolution of heatmaps is not enough to estimate keypoint coordinates from high-compressed ones. To solve this problem, CPN \cite{chen2018cascadedCPN} combined all levels of feature representations using a cascaded pyramid network. Similarly, HRNet \cite{sun2019deephrnet} repeated high-to-low and low-to-high feature fusions to maintain high-resolution representations. HigherHRNet \cite{cheng2020higherhrnet} reconstructed HRNet with an efficient architecture by reducing its complexity. Despite of aforementioned progresses, it is still difficult to handle dense and overlapped keypoints. 

\begin{figure*}[t]
    \centering
    \includegraphics[width=1.0\textwidth]{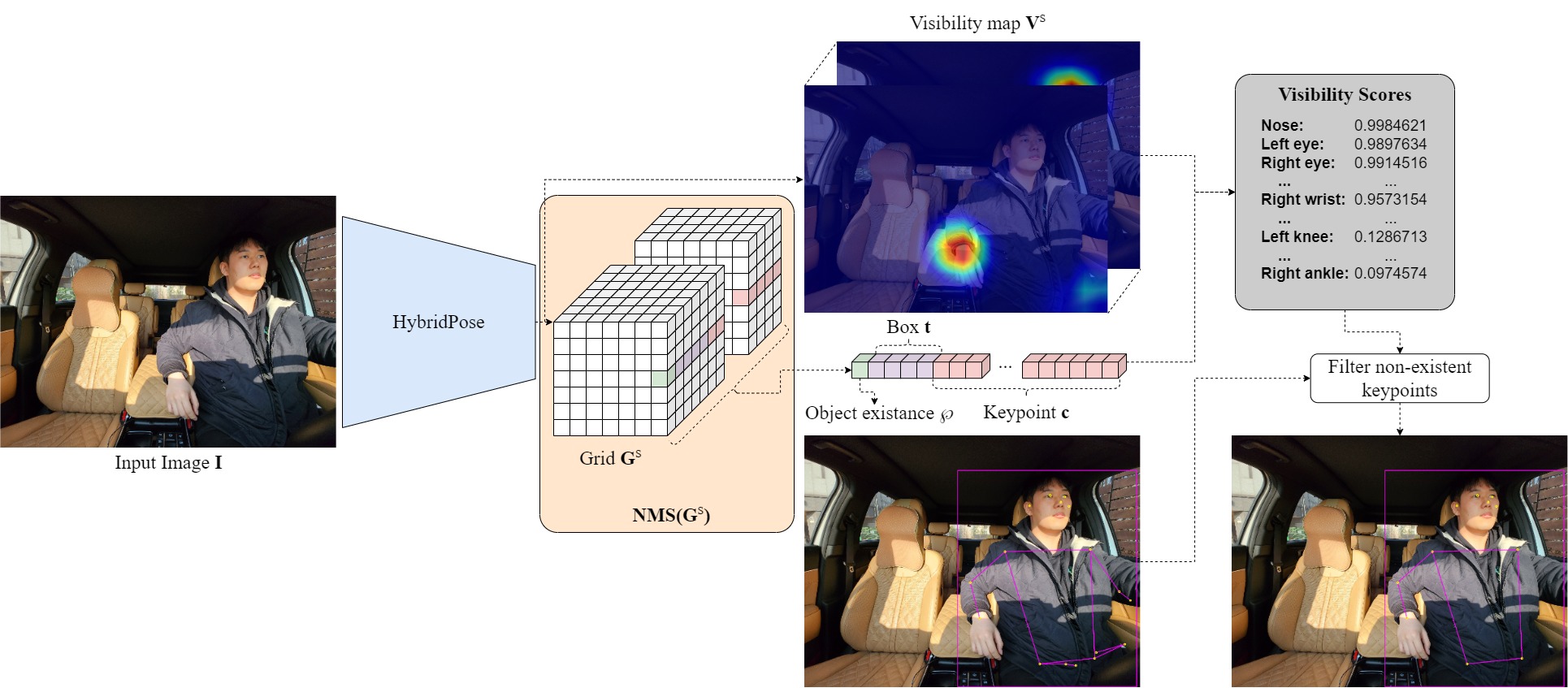}
    \caption{Overall architecture of the proposed HybridPose. The visibility maps facilitate HybridPose to filter non-existent keypoints out using their visibility scores.}
    \label{architecture}
\end{figure*}

\section{Proposed Method}
\label{sec:method}
We aim to overcome drawbacks of regression and heatmap-based methods by maximizing their strengths. To implement this concept, we first introduce an architecture of HybridPose that simultaneously estimates human bounding boxes, keypoint coordinates, and visibility maps.
Then, we describe the self-correlation loss to adjust probability distributions of visibility maps along with their corresponding regression results.

\subsection{Architecture}
HybridPose is comprised of three parts in terms of its output; human bounding boxes, keypoint coordinates, and visibility maps as illustrated on Fig. \ref{architecture}. To yields these predictions, HybridPose adopts a YOLO-style deep neural network which is robust in multi-resolution feature extraction. Specifically, HybridPose analyzes an input image $\textbf{I} \in \mathbb{R}^{h\times w\times 3}$ using the YOLO-style network, and yields two kinds of grids sets respectively. One is a set of grids for the keypoints, and the other is for the visibility maps. The set of grids for keypoints is $\textbf{G}^s \in \mathbb{R}^{\frac{h}{s}\times \frac{w}{s}\times N_{a} \times N_{o}}$, where $N_a$ and $N_o$ are the number of anchor and output channels respectively. In addition, a set of the visibility maps is $\textbf{V}^s \in \mathbb{R}^{\frac{h}{s}\times \frac{w}{s} \times N_a \times K}$, where $K$ is the number of keypoints. Then, an image patch $\textbf{I}_p = \textbf{I}_{si:s(i+1),sj:s(j+1)}$ is compressed into two kinds of the grid cells both $\textbf{G}^{s}_{ij}$ and $\textbf{V}^{s}_{ij}$, where $s \in \{8, 16, 32, 64\}$ is a set of scale factors. The scale factor contains diverse receptive fields to detect different scales of objects, i.e., detecting small-sized objects with lower scale factors and vice versa. In addition, HybridPose is capable of detecting different object shapes by utilizing multiple anchor boxes $\textbf{A}^s = \{(A_{w_a}, A_{h_a})\}^{N_a}_{a=1}$, where $a$ is an index of anchors.

Each grid cell $\textbf{G}^{s}_{i,j,a}$ contains the probability of object existence $\wp$, human bounding boxes $\textbf{t}=(t_x, t_y, t_w, t_h)$, keypoint coordinates $\textbf{c}=\{(c_{xk},c_{yk})\}^{K}_{k=1}$. Therefore, the number of the output channels for a grid cell $\textbf{G}^{s}_{i,j,a}$ is $5+2K$. Following previous works \cite{wang2021scaledyolov4, https://doi.org/10.5281/zenodo.7347926yolov5, mcnally2022rethinkingkapao}, coordinates of a human bounding box $\textbf{t}$ are positioned on the grid space, thus these are relative to the origin of a grid cell $(i.j)$:

\begin{equation}
    t_x', t_y' = (2\sigma(t_x)-0.5, 2\sigma(t_y)-0.5)
\end{equation}
\begin{equation}
    t_w', t_h'=(\frac{A_w}{s}(2\sigma(t_w))^2, \frac{A_h}{s}(2\sigma(t_h))^2),
\end{equation}
where $\sigma$ is the sigmoid function. Similar to the human bounding box, the keypoint coordinates are positioned on the grid space:

\begin{equation}
    c_{xk}', c_{yk}'=(\frac{A_w}{s}(4\sigma(c_{xk})-2), \frac{A_h}{s}(4\sigma(c_{yk})-2)).
\end{equation}

As mentioned in Section \ref{sec:relate}, the conventional regression-based methods cause undesirable results with invisible keypoints in a scene since it is limited to determine the keypoint uncertainty in terms of their visibility. To this end, HybridPose utilizes visibility maps which are a set of probability distributions for each keypoint. In detail, shallow convolutional layers analyze feature maps $f^s_D \in \mathbb{R}^{\frac{h}{s}\times \frac{w}{s}\times N_c}$ from a decoder part $D$ of HybridPose:

\begin{equation}
    \textbf{V}^s = \phi(f^s_D),
\end{equation}
where $\phi$ and $N_c$ are the shallow convolutional layers and the number of feature channels respectively. Then, the visibility of a specific keypoint can be defined as follows:
\begin{equation}
    {v}_{c_{xk}',c_{yk}',a,k}=\mathbf{V}^{s}(c_{xk},c_{yk},a,k),
    \label{visscore}
\end{equation}
where $v_{c_{xk}',c_{yk}',a,k}$ is the $k$-th keypoint visibility positioned $(c_{xk}',c_{yk}')$ in a scene. If a keypoint visibility value is larger than a certain threshold, it can be considered as visible in the scene. In the opposite case, the keypoints can be considered as invisible.
Therefore, HybridPose is capable of covering the uncertainty of keypoint existence by adopting visibility maps as shown in Fig. \ref{architecture}.

\begin{figure*}
    \centering
    \includegraphics[width=0.95\textwidth]{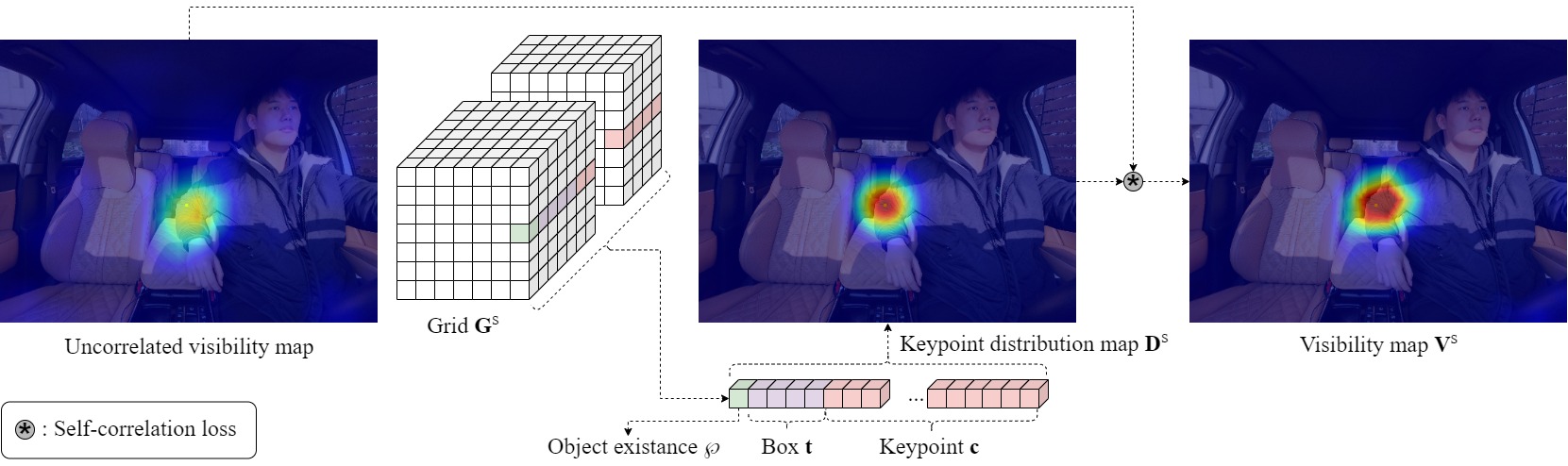}
    \caption{Flow of the proposed self-correlation loss. }
    \label{selfcorr}
\end{figure*}

\subsection{Self-Correlation Loss}
HybridPose predicts keypoint coordinates and their existence independently. Thus, the locations of the highest activation in a visibility map are uncorrelated with coordinates of its corresponding keypoint. In this case, the visibility map cannot be ensured the reliability of determining keypoint existence. To tackle these issue, the self-correlation loss inject spatial relationships between keypoint coordinates and visibility maps.

To compare both them in the spatial domain, the keypoint coordinates $\textbf{c}_{i,j,a}$ in a grid cell $\textbf{G}^{s}_{i,j,a}$ are converted into a keypoint distribution map $\textbf{d}_{u,v,a} \in \mathbb{R}^{\frac{h}{s}\times \frac{w}{s} \times K}$ as follows:
\begin{equation}
    \begin{aligned}
        \textbf{d}_{u,v,a} & = \Psi(\textbf{c}_{i,j,a}) \\
                           & = \wp e^{-((u-c_{xk})^2+(v-c_{yk})^2)/2\sigma^2} \\ 
                           & s.t. \quad \left \| u-c_{xk} \right \| \leq 3\sigma \quad \left \| v-c_{yk} \right \| \leq 3\sigma,
    \end{aligned}
    \label{distribution}
\end{equation}
where $\Psi(\cdot)$ is a conversion function, and $\sigma$ denotes the standard deviation. In addition, the probability of object existence $\wp$ is multiplied when obtaining the keypoint distribution map $\textbf{d}_{u,v,a}$ to consider objectness in its corresponding grid cell. Eq. \ref{distribution} is repeated in each grid cell $i\times j$ to combine all outputs. Following this step, a set of keypoint distribution maps $\textbf{D}^s \in \mathbb{R}^{\frac{h}{s}\times \frac{w}{s}\times N_{a}\times K}$ are described as follows:
\begin{equation}
    \textbf{D}^s = \{\max[\textbf{d}_{0,0,a}, \textbf{d}_{0,1,a}, ..., \textbf{d}_{i,j,a}]\}^{N_a}_{a=1},
    \label{max}
\end{equation}
where $\max$ is pixel-wisely operated to filter out lower activations of dense keypoints in the same class.

Finally, self-correlation loss is applied to a set of visibility maps and keypoint distribution maps as follows:
\begin{equation}
    \mathcal{L}_{corr} = 1 - (\frac{\sum ((D^s - \bar{D^s})*(V^s - \bar{V^s}))}{\sqrt{\sum (D^s - \bar{D^s})^2}*\sqrt{\sum (V^s - \bar{V^s})^2}+\epsilon})^2 ,
\end{equation}
where $\bar{(\cdot)}$ denotes arithmetic mean.

\subsection{Learning Objectives}
In addition to the loss functions discussed so far, we apply a couple of more loss functions as follows:

\textit{Objectiveness}: Let $\wp$ and $\hat{\wp}$ be the predicted probability of object existence and its ground-truth, then an objectiveness loss is described as:

\begin{equation}
\begin{aligned}
    \mathcal{L}_{obj} = \sum_{s}\frac{\omega_s}{n(\textbf{G}^s)}\sum_{\textbf{G}^s} & -(\wp \log(\hat{\wp}* \text{IoU} (\textbf{t},\hat{\textbf{t}})) \\
    &+(1-\wp)\log (1-\hat{\wp}*\text{IoU}(\textbf{t},\hat{\textbf{t}}))),
\end{aligned}
\end{equation}
where $\omega_s$ denotes a set of grid weight, $n(\cdot)$ is the number of its instance. Similar to YOLO \cite{redmon2016youyolo}, ground-truth of objectiveness $\hat{\wp}$ is multiplied with intersection over union (IoU).

\textit{Human bounding box and pose estimation}: Let $\hat{O}$ and $\hat{O^p}$ denote the target object and its corresponding pose, then loss functions to estimate human bounding boxes and keypoint coordinates are described as:

\begin{equation}
    \mathcal{L}_{box} = \sum_{s} \frac{1}{n(\hat{O} \in \textbf{G}^s)} \sum_{\hat{O} \in \textbf{G}^s} (1- \text{IoU}(\textbf{t},\hat{\textbf{t}}))
\end{equation}

\begin{equation}
    \mathcal{L}_{pose} = \sum_{s} \frac{1}{n(\hat{O^p} \in \textbf{G}^s)} \sum_{\hat{O^p} \in \textbf{G}^s} \sum_{k=1}^{K} \delta_k \left \| \textbf{c}_k - \hat{\textbf{c}_k} \right \| _2,
\end{equation}
where $\delta_k$ is set to $1$ if a $k$-th keypoint exists and otherwise $0$.

\textit{Visibility map}: Let $\hat{\textbf{c}}$ denotes ground-truth of keypoint coordinates, then the target visibility map $\hat{\textbf{V}^s}$ is generated using Eq. \ref{distribution} and \ref{max} without the probability of object existence $\wp$. After that, a loss function to predict the visibility maps is described as:
\begin{equation}
    \mathcal{L}_{vis} = \sum_{s} \frac{1}{n(\textbf{V}^s)} \sum_{\textbf{V}^s} \left \| \textbf{V}^s - \hat{\textbf{V}^s} \right \|_2
\end{equation}

\textit{Total loss}: we aggregate all loss functions to train HybridPose. Thus, the total loss can be defined as:

\begin{equation}
    \mathcal{L}_{total} = \alpha \mathcal{L}_{obj} + \beta \mathcal{L}_{box} + \gamma \mathcal{L}_{pose} + \zeta \mathcal{L}_{vis}  + \lambda \mathcal{L}_{corr},
\end{equation}
where $\alpha$, $\beta$, $\gamma$, $\zeta$, and $\lambda$ are weighting factors for their corresponding loss functions.

\begin{figure*}[t]
    \centering
    \includegraphics[width=1.0\textwidth]{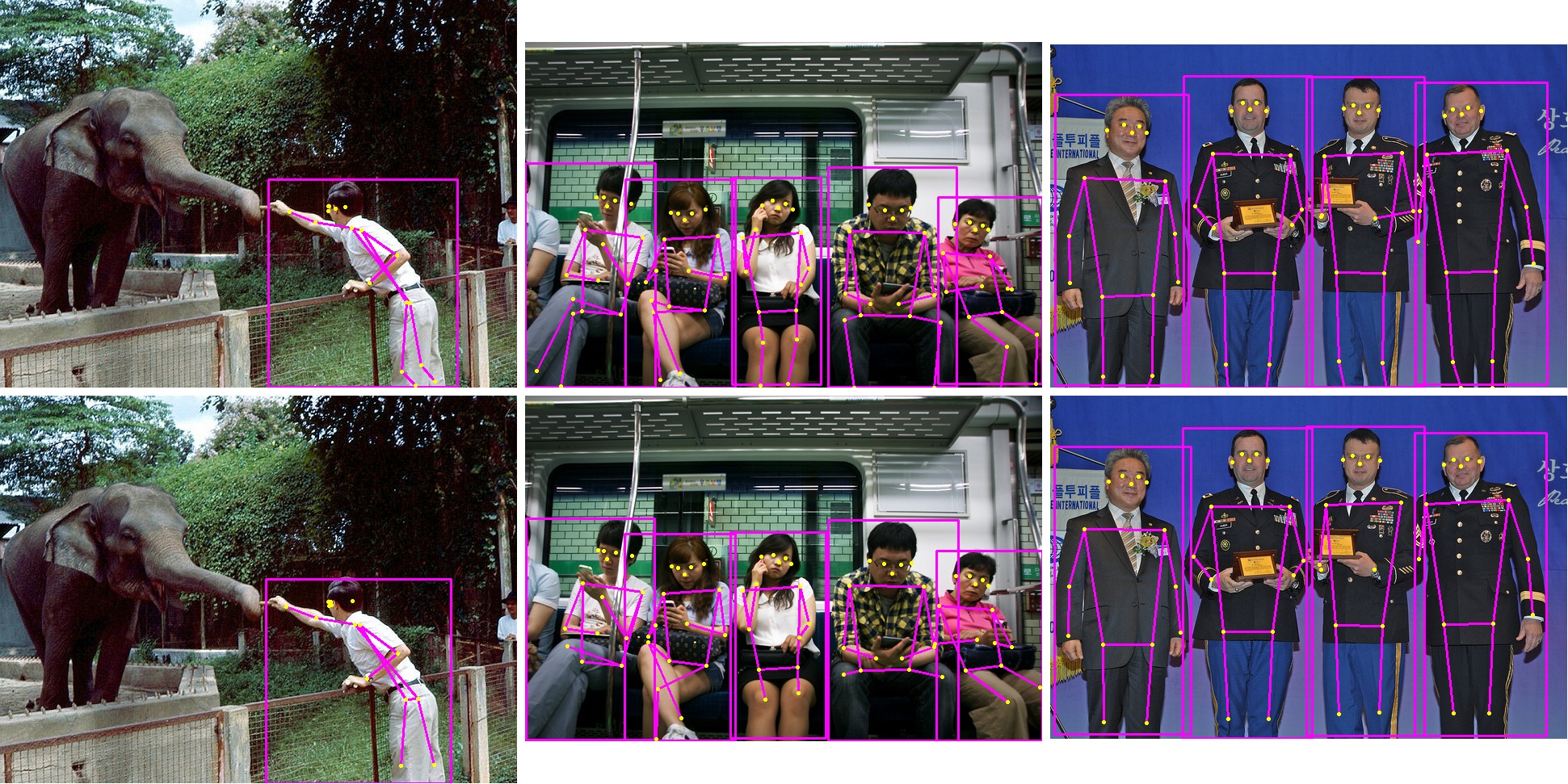}
    \caption{Examples of human pose estimation on \textbf{COCO2017 val}. Figures in the first row are estimated by KAPAO-S \cite{mcnally2022rethinkingkapao}, and second row ones are HybridPose-S results.}
    \label{vis_comp}
\end{figure*}

\begin{table*}[t]
\centering
\caption{Comparisons with existing single-stage methods on \textbf{COCO2017 val} split. Inference time is measured on a single Titan X GPU. MI and PP denote model inference and post-processing respectively. TTA (Test Time Augmentation) is not applied in this experiment.}
\begin{tabular}{l||ccc||cccccc}
\hline
\multirow{2}{*}{Method} & \multicolumn{3}{c||}{Speed}                                             & \multicolumn{6}{c}{Accuracy}                                                                                                                    \\ \cline{2-10} 
                        & \multicolumn{1}{c|}{Input Size} & \multicolumn{1}{c|}{MI (ms)}      & PP (ms)    & \multicolumn{1}{c|}{AP}   & \multicolumn{1}{c|}{$\text{AP}^{.50}$} & \multicolumn{1}{c|}{$\text{AP}^{.75}$} & \multicolumn{1}{c|}{$\text{AP}^{M}$}  & \multicolumn{1}{c|}{$\text{AP}^{L}$}  & AR   \\ \hline \hline
SWAHR-W48 \cite{luo2021rethinkingswahr}                  & \multicolumn{1}{c|}{640}        & \multicolumn{1}{c|}{65.9}  & 133.6 & \multicolumn{1}{c|}{67.3} & \multicolumn{1}{c|}{87.1} & \multicolumn{1}{c|}{72.9} & \multicolumn{1}{c|}{62.1} & \multicolumn{1}{c|}{75.0} & 73.0 \\ \hline
DEKR-W32 \cite{geng2021bottomdekr}                & \multicolumn{1}{c|}{640}        & \multicolumn{1}{c|}{193.3}  & 4.95 & \multicolumn{1}{c|}{67.2} & \multicolumn{1}{c|}{86.3} & \multicolumn{1}{c|}{73.7} & \multicolumn{1}{c|}{61.7} & \multicolumn{1}{c|}{77.0} & 73.0 \\ \hline
DEKR-W48 \cite{geng2021bottomdekr}                & \multicolumn{1}{c|}{640}        & \multicolumn{1}{c|}{381.8}  & 5.07 & \multicolumn{1}{c|}{70.2} & \multicolumn{1}{c|}{87.8} & \multicolumn{1}{c|}{76.8} & \multicolumn{1}{c|}{66.2} & \multicolumn{1}{c|}{77.9} & 75.9 \\ \hline
KAPAO-S \cite{mcnally2022rethinkingkapao}                & \multicolumn{1}{c|}{1280}       & \multicolumn{1}{c|}{30.8}  & 3.57 & \multicolumn{1}{c|}{63.0} & \multicolumn{1}{c|}{86.3} & \multicolumn{1}{c|}{69.5} & \multicolumn{1}{c|}{58.0} & \multicolumn{1}{c|}{70.8} & 70.2 \\ \hline
KAPAO-M \cite{mcnally2022rethinkingkapao}                 & \multicolumn{1}{c|}{1280}       & \multicolumn{1}{c|}{64.6}  & 3.58 & \multicolumn{1}{c|}{68.5} & \multicolumn{1}{c|}{88.5} & \multicolumn{1}{c|}{75.0} & \multicolumn{1}{c|}{63.8} & \multicolumn{1}{c|}{76.3} & 75.5 \\ \hline
KAPAO-L \cite{mcnally2022rethinkingkapao}                 & \multicolumn{1}{c|}{1280}       & \multicolumn{1}{c|}{110.9} & 3.87 & \multicolumn{1}{c|}{70.6} & \multicolumn{1}{c|}{89.6} & \multicolumn{1}{c|}{77.1} & \multicolumn{1}{c|}{66.4} & \multicolumn{1}{c|}{\textbf{77.5}} & \textbf{77.4} \\ \hline \hline
HybridPose-S            & \multicolumn{1}{c|}{1280}       & \multicolumn{1}{c|}{36.1}  & 1.61 & \multicolumn{1}{c|}{63.0} & \multicolumn{1}{c|}{85.9} & \multicolumn{1}{c|}{69.7} & \multicolumn{1}{c|}{59.5} & \multicolumn{1}{c|}{69.7} & 69.9 \\ \hline
HybridPose-M            & \multicolumn{1}{c|}{1280}       & \multicolumn{1}{c|}{65.7}  & 1.49 & \multicolumn{1}{c|}{68.7} & \multicolumn{1}{c|}{88.7} & \multicolumn{1}{c|}{75.9} & \multicolumn{1}{c|}{65.2} & \multicolumn{1}{c|}{75.6} & 75.4 \\ \hline
HybridPose-L            & \multicolumn{1}{c|}{1280}       & \multicolumn{1}{c|}{112.1} & 1.54 & \multicolumn{1}{c|}{\textbf{70.7}} & \multicolumn{1}{c|}{\textbf{89.8}} & \multicolumn{1}{c|}{\textbf{77.6}} & \multicolumn{1}{c|}{\textbf{68.1}} & \multicolumn{1}{c|}{76.6} & \textbf{77.4} \\ \hline
\end{tabular}
\label{cocoval}
\end{table*}

\section{Experimental Results}
\label{sec:experiment}

\subsection{Experimental Setting}
\textbf{Dataset}.
We adopt a COCO \cite{lin2014microsoft} keypoints detection dataset, which is a challenging and widely adopted dataset for human pose estimation. COCO is split into $train$, $val$, and $test$-$dev$ respectively, and it contains over 200K images. In addition, each person is labeled with human bounding boxes and their corresponding 17 keypoints. To evaluate the performance in terms of estimation accuracy, we adopt the standard average precision (AP) and recall (AR) by utilizing object keypoint similarity (OKS). To be specific, AP includes $\text{AP}^{.50}$ (AP at $OKS=0.50$), $\text{AP}^{.75}$, AP (mean of AP scores from
$OKS=0.50$ to $OKS=0.95$ with the increment as 0.05), $\text{AP}^{M}$ (AP scores for person of medium sizes), and $\text{AP}^{L}$ (AP scores for person of large sizes). Moreover, HybridPose is trained and evaluated on CrowdPose \cite{li2019crowdpose} which is split into $trainval$ and $test$. Each split contains 12K and 8K images respectively. CrowdPose divides test scenes into three crowding levels; $\text{AP}^{E}$ (mean of AP scores from $OKS=0$ to $OKS=0.1$), $\text{AP}^{M}$ (mean of AP scores from $OKS=0.1$ to $OKS=0.8$), $\text{AP}^{H}$ (mean of AP scores from $OKS=0.8$ to $OKS=1.0$). In addition, the algorithm performance in terms of execution times has been reported on both a Titan X GPU and an edge device.

\begin{figure*}
    \centering
    \includegraphics[width=1.0\textwidth]{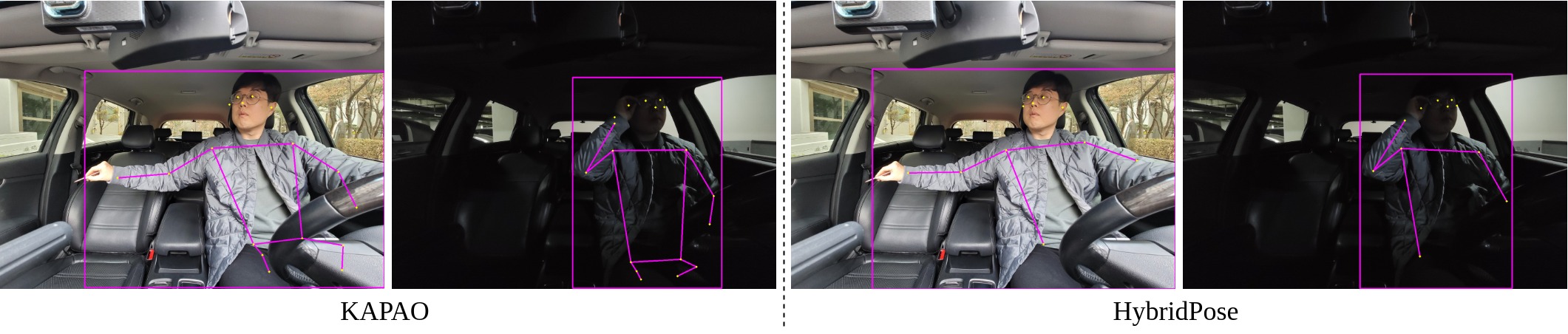}
    \caption{Examples of human pose estimation on general DMS (Driver Monitoring System). First two figures are obtained by KAPAO-S \cite{mcnally2022rethinkingkapao}, and others are estimated by HybridPose-S}
    \label{general}
\end{figure*}

\begin{table}[t]
\scriptsize
\centering
\caption{Comparisons with state-of-the-art methods on \textbf{COCO2017 \textit{test-dev}} split. TTA (Test Time Augmentation) is applied in this experiment.}
\begin{tabular}{lcccccc}
\hline
\multicolumn{1}{l||}{Method}          & \multicolumn{1}{c|}{AP}   & \multicolumn{1}{c|}{$\text{AP}^{.50}$} & \multicolumn{1}{c|}{$\text{AP}^{.75}$} & \multicolumn{1}{c|}{$\text{AP}^{M}$} & \multicolumn{1}{c|}{$\text{AP}^{L}$} & \multicolumn{1}{c}{AR} \\ \hline \hline
\multicolumn{7}{c}{Two-stage methods}                                                                                                                                                                                                                                                                  \\ \hline \hline
\multicolumn{1}{l||}{Mask-RCNN \cite{he2017mask}}       & \multicolumn{1}{c|}{63.1} & \multicolumn{1}{c|}{87.3}                        & \multicolumn{1}{c|}{68.7}                        & \multicolumn{1}{c|}{57.8}                  & \multicolumn{1}{c|}{71.4}                  & \multicolumn{1}{c}{-}         \\ \hline
\multicolumn{1}{l||}{CPN \cite{chen2018cascadedCPN}}             & \multicolumn{1}{c|}{72.1} & \multicolumn{1}{c|}{91.4}                        & \multicolumn{1}{c|}{80.0}                        & \multicolumn{1}{c|}{68.7}                  & \multicolumn{1}{c|}{77.2}                  & \multicolumn{1}{c}{78.5}        \\ \hline
\multicolumn{1}{l||}{SimpleBaseline \cite{xiao2018simple}}  & \multicolumn{1}{c|}{73.7} & \multicolumn{1}{c|}{91.9}                        & \multicolumn{1}{c|}{81.1}                        & \multicolumn{1}{c|}{70.3}                  & \multicolumn{1}{c|}{80.0}                  & \multicolumn{1}{c}{79.0}        \\ \hline
\multicolumn{1}{l||}{HRNet-W48 \cite{sun2019deephrnet}}       & \multicolumn{1}{c|}{75.5} & \multicolumn{1}{c|}{92.5}                        & \multicolumn{1}{c|}{83.3}                        & \multicolumn{1}{c|}{71.9}                  & \multicolumn{1}{c|}{81.5}                  & \multicolumn{1}{c}{80.5}         \\ \hline
\multicolumn{1}{l||}{RLE \cite{li2021humanRLE}}             & \multicolumn{1}{c|}{75.7} & \multicolumn{1}{c|}{92.3}                        & \multicolumn{1}{c|}{82.9}                        & \multicolumn{1}{c|}{72.3}                  & \multicolumn{1}{c|}{81.3}                  & \multicolumn{1}{c}{-}        \\ \hline \hline
\multicolumn{7}{c}{Single-stage methods}                                                                                                                                                                                                                                                               \\ \hline \hline
\multicolumn{1}{l||}{OpenPose \cite{cao2017realtime}}        & \multicolumn{1}{c|}{61.8} & \multicolumn{1}{c|}{84.9}                        & \multicolumn{1}{c|}{67.5}                        & \multicolumn{1}{c|}{57.1}                  & \multicolumn{1}{c|}{68.2}                  & \multicolumn{1}{c}{66.5}           \\ \hline
\multicolumn{1}{l||}{CenterNet \cite{zhou2019objects}}       & \multicolumn{1}{c|}{63.0} & \multicolumn{1}{c|}{86.8}                        & \multicolumn{1}{c|}{69.6}                        & \multicolumn{1}{c|}{58.9}                  & \multicolumn{1}{c|}{70.4}                  & \multicolumn{1}{c}{-}              \\ \hline
\multicolumn{1}{l||}{HigherHRNet \cite{cheng2020higherhrnet}} & \multicolumn{1}{c|}{70.5} & \multicolumn{1}{c|}{89.3}                        & \multicolumn{1}{c|}{77.2}                        & \multicolumn{1}{c|}{66.6}                  & \multicolumn{1}{c|}{75.8}                  & \multicolumn{1}{c}{74.9}           \\ \hline
\multicolumn{1}{l||}{\quad +SWAHR \cite{luo2021rethinkingswahr}}          & \multicolumn{1}{c|}{\textbf{72.0}} & \multicolumn{1}{c|}{90.7}                        & \multicolumn{1}{c|}{\textbf{78.8}}                        & \multicolumn{1}{c|}{67.8}                  & \multicolumn{1}{c|}{\textbf{77.7}}                  & \multicolumn{1}{c}{-}              \\ \hline
\multicolumn{1}{l||}{DEKR-W48 \cite{geng2021bottomdekr}}        & \multicolumn{1}{c|}{71.0} & \multicolumn{1}{c|}{89.2}                        & \multicolumn{1}{c|}{78.0}                        & \multicolumn{1}{c|}{67.1}                  & \multicolumn{1}{c|}{76.9}                  & \multicolumn{1}{c}{76.7}           \\ \hline
\multicolumn{1}{l||}{CenterGroup \cite{braso2021center}} & \multicolumn{1}{c|}{71.4} & \multicolumn{1}{c|}{90.5}                        & \multicolumn{1}{c|}{78.1}                        & \multicolumn{1}{c|}{67.2}                  & \multicolumn{1}{c|}{77.5}                  & \multicolumn{1}{c}{-}              \\ \hline
\multicolumn{1}{l||}{KAPAO-S \cite{mcnally2022rethinkingkapao}}         & \multicolumn{1}{c|}{63.8} & \multicolumn{1}{c|}{88.4}                        & \multicolumn{1}{c|}{70.4}                        & \multicolumn{1}{c|}{58.6}                  & \multicolumn{1}{c|}{71.7}                  & \multicolumn{1}{c}{71.2}           \\ \hline
\multicolumn{1}{l||}{KAPAO-M \cite{mcnally2022rethinkingkapao}}         & \multicolumn{1}{c|}{68.8} & \multicolumn{1}{c|}{90.5}                        & \multicolumn{1}{c|}{76.5}                        & \multicolumn{1}{c|}{64.3}                  & \multicolumn{1}{c|}{76.0}                  & \multicolumn{1}{c}{76.3}           \\ \hline
\multicolumn{1}{l||}{KAPAO-L \cite{mcnally2022rethinkingkapao}}         & \multicolumn{1}{c|}{70.3} & \multicolumn{1}{c|}{\textbf{91.2}}                        & \multicolumn{1}{c|}{77.8}                        & \multicolumn{1}{c|}{66.3}                  & \multicolumn{1}{c|}{76.8}                  & \multicolumn{1}{c}{77.7}           \\ \hline \hline
\multicolumn{1}{l||}{HybridPose-S}    & \multicolumn{1}{c|}{63.1}     & \multicolumn{1}{c|}{87.8}                            & \multicolumn{1}{c|}{70.2}                            & \multicolumn{1}{c|}{60.1}                      & \multicolumn{1}{c|}{69.7}                      & \multicolumn{1}{c}{71.1}               \\ \hline
\multicolumn{1}{l||}{HybridPose-M}    & \multicolumn{1}{c|}{67.4}     & \multicolumn{1}{c|}{89.8}                            & \multicolumn{1}{c|}{75.5}                            & \multicolumn{1}{c|}{65.0}                      & \multicolumn{1}{c|}{73.3}                      & \multicolumn{1}{c}{75.8}               \\ \hline
\multicolumn{1}{l||}{HybridPose-L}    & \multicolumn{1}{c|}{69.8}     & \multicolumn{1}{c|}{90.7}                            & \multicolumn{1}{c|}{78.0}                            & \multicolumn{1}{c|}{\textbf{68.0}}                      & \multicolumn{1}{c|}{75.1}                      & \multicolumn{1}{c}{\textbf{77.8}}               \\ \hline
\end{tabular}
\label{cocotest}
\end{table}

\begin{table}[t]
\scriptsize
\centering
\caption{Comparisons with single-stage state-of-the-art methods on \textbf{CrowdPose} split. TTA (Test Time Augmentation) is applied in this experiment.}
\begin{tabular}{l||c|c|c|c|c|c}
\hline
Method                                      & AP   & $\text{AP}^{.50}$ & $\text{AP}^{.75}$ & $\text{AP}^{E}$  & $\text{AP}^{M}$  & $\text{AP}^{H}$  \\ \hline \hline
OpenPose \cite{cao2017realtime}             & -    & -    & -    & 62.7 & 48.7 & 32.3 \\ \hline
HigherHRNet \cite{cheng2020higherhrnet} & 67.6 & 87.4 & 72.6 & 75.8 & 68.1 & 58.9 \\ \hline
DEKR-W48 \cite{geng2021bottomdekr}          & 68.0 & 85.5 & 73.4 & 76.6 & 68.8 & 58.4 \\ \hline
CenterGroup \cite{braso2021center}      & 70.0 & 88.9 & 75.7 & 77.7 & 70.8 & \textbf{63.2} \\ \hline
KAPAO-S \cite{mcnally2022rethinkingkapao}   & 63.8 & 87.7 & 69.4 & 72.1 & 64.8 & 53.2 \\ \hline
KAPAO-M \cite{mcnally2022rethinkingkapao}   & 67.1 & 88.8 & 73.4 & 75.2 & 68.1 & 56.9 \\ \hline
KAPAO-L \cite{mcnally2022rethinkingkapao}   & 68.9 & 89.4 & 75.6 & 76.6 & 69.9 & 59.5 \\ \hline \hline
HybridPose-S                                & 63.9 & 88.4 & 71.2 & 73.5 & 65.1 & 52.0 \\ \hline
HybridPose-M                                & 67.4 & 89.6 & 75.0 & 76.2 & 68.6 & 55.5 \\ \hline
HybridPose-L                                & \textbf{70.3} & \textbf{90.3} & \textbf{78.6} & \textbf{78.5} & \textbf{71.7} & 58.8 \\ \hline
\end{tabular}
\label{crowdpose}
\end{table}

\textbf{Implementation Details}.
Following an existing MPPE method \cite{mcnally2022rethinkingkapao}, HybridPose adopts both YOLO backbone \cite{https://doi.org/10.5281/zenodo.7347926yolov5} and feature pyramid network (FPN) \cite{lin2017feature}. To train HybridPose on COCO and CrowdPose, an initial learning rate is set to 0.0001, and 0.5 is multiplied if AP is unreached to the best performance during 20 epochs. In addition, the both mosaic and flip are adopted as data augmentation \cite{bochkovskiy2020yolov4}. HybridPose is optimized with stochastic gradient descent (SGD) over 8 NVIDIA V100 GPUs, and the batch sizes of HybridPose-L/M/S are 48/72/128. On COCO2017 $val$, $test$-$dev$ splits, and CrowdPose, the proposed HybridPose is performed both subjective and objective quality evaluations.

\begin{figure*}
    \centering
    \includegraphics[width=1.0\textwidth]{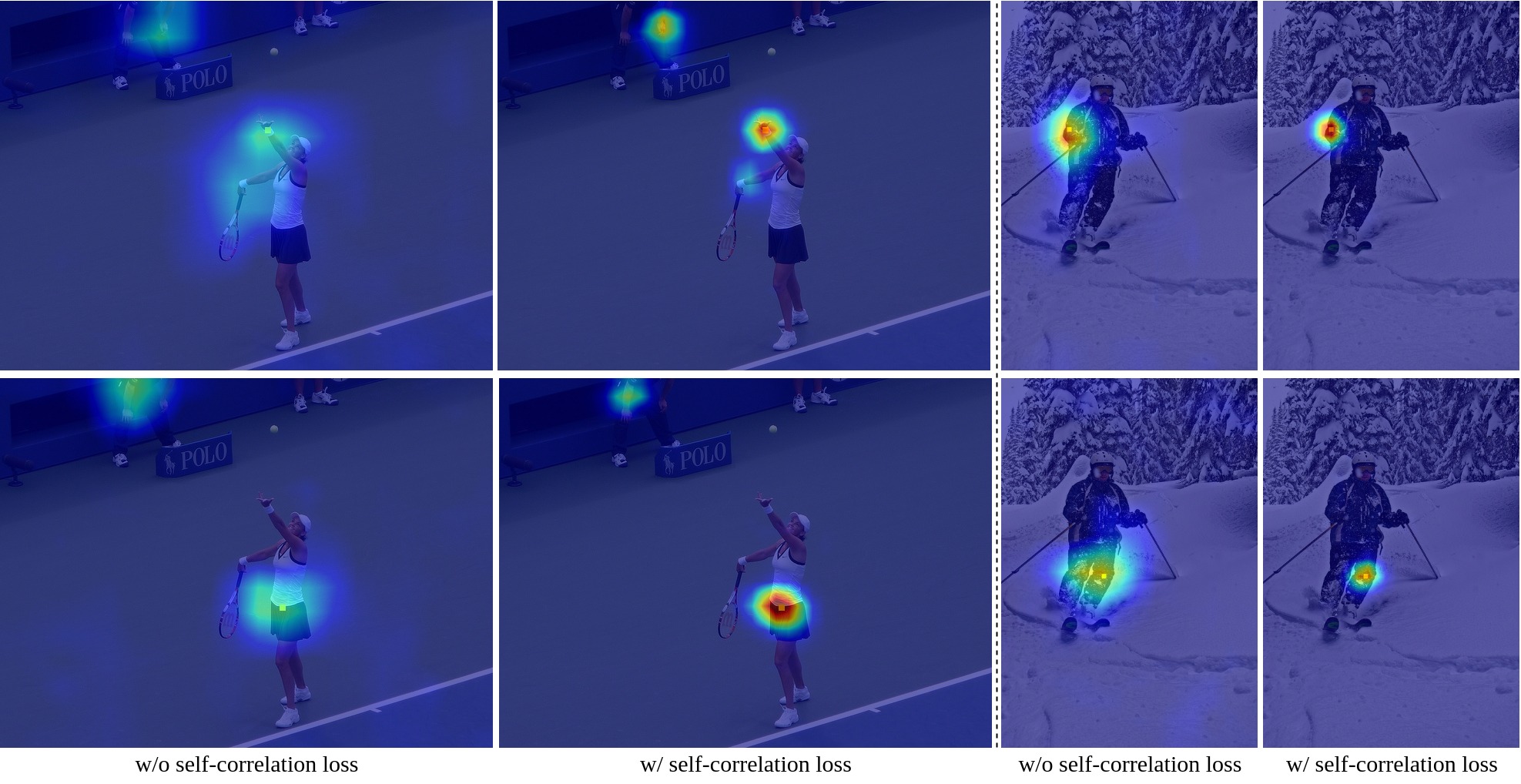}
    \caption{Effects of self-correlation loss on visual quality of visibility maps.}
    \label{corr_comp}
\end{figure*}

\subsection{Analysis for Subjective and Objective Quality}
We analysis HybridPose in terms of both subjective and objective qualities to validate our contributions. Furthermore, we provide qualitative and quantitative comparisons with state-of-the-art methods on COCO and CrowdPose.

For evaluating subjective quality, we provide visual examples of human pose estimation on COCO. As shown in the first row of Fig. \ref{vis_comp}, KAPAO \cite{mcnally2022rethinkingkapao}, which is the state-of-the-art method among existing single-stage methods, yields wrong placed keypoints in ankles. On the other hand, HybridPose display only valid keypoints by filtering invisible keypoints with lower visibility scores as described in the second row of Fig. \ref{vis_comp}. Furthermore, HybridPose is capable of detecting keypoints of multi-person in a single step while considering their visibility.

For objective quality, Table \ref{cocoval} summarizes performance comparisons between HybridPose and state-of-the-art methods in terms of the pose estimation accuracy on a COCO2017 $val$ split. In this comparison, test time augmentation is not applied. As described in Table \ref{cocoval}, HybridPose slightly outperforms existing single-stage pose estimation algorithms on COCO2017 $val$. Moreover, inference time of HybridPose is faster than KAPAO \cite{mcnally2022rethinkingkapao} although both methods share similar network architectures. To be specific, HybridPose adopts additional shallow convolutional layers to predict visibility maps. Thus, these layers cause increase in model inference time. On the other hand, post-processing time is declined in HybridPose compared with KAPAO since post-processing is much simplified by eliminating keypoint refinement. In addition, we provide Table \ref{cocotest} to compare HybridPose on a COCO2017 $test$-$dev$ split with existing two-stage and single-stage methods. HybridPose achieves the best performance on AR with 77.8 while AP is 2.2 lower than SWAHR \cite{luo2021rethinkingswahr}. However, inference speed of HybridPose is around 1.8 times faster than SWAHR as described in Table \ref{cocoval}. Moreover, HybridPose shows state-of-the-art AP in detecting medium-sized persons. We further provide comparisons of body pose estimation performance on CrowdPose with existing single-stage methods. As described in Table \ref{crowdpose}, HybridPose achieves the state-of-the-art performance on AP with 70.3. To be specific, HybridPose outperforms existing methods in AP of two crowding levels by showing 78.5 and 71.7 in $\text{AP}^{E}$ and $\text{AP}^{M}$ respectively. Those experimental results indicate that HybridPose is more robust in detecting keypoints than state-of-the-art methods.

Consequently, visibility maps facilitate HybridPose to eliminate invisible keypoints in a scene while achieving single-stage pose estimation without significant degradation of performance.

\begin{figure}[t]
    \centering
    \includegraphics[width=0.5\textwidth]{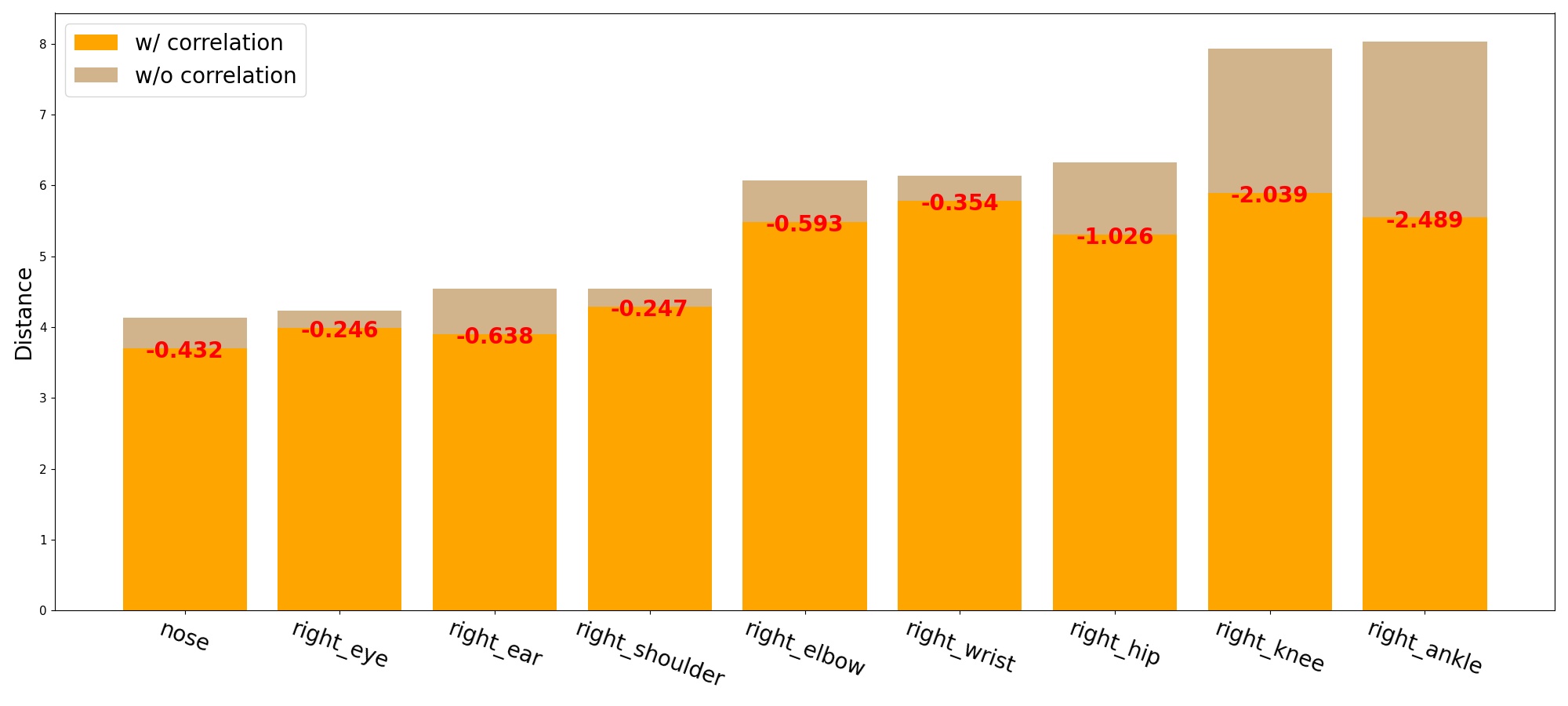}
    \caption{Ablation study of self-correlation loss on the distance of peak values.}
    \label{vis_dist}
\end{figure}

\subsection{Generalizability}
For generalizability experiment, we adopt example images taken from the driver monitoring system (DMS), which is widely adopted general experimental scenario and required the MPPE algorithms. In the DMS scenario, some keypoints are not only occluded by wheel but also located on the out of the images. It can be observed from the Fig. \ref{general} that KAPAO \cite{mcnally2022rethinkingkapao} cannot handle invisible keypoints effectively and produce incorrect pose estimations. However, HybridPose is capable of estimating only visible keypoints accurately while filtering out invisible ones. This experimental result demonstrates that the proposed HybridPose can achieve the effective estimation for both human pose and its visibility in any arbitrary scene. 

In addition, we optimize HybridPose-S to conduct human pose estimation on a low-power device. To implement it, we resize input resolution for HybridPose-S as $320\times 320$, and adopt width multiplier in every convolutional layer to reduce their channels. Moreover, python-based codes are converted into C++ to conduct HybridPose-S on the low-power device. Therefore, HybridPose-S achieves the real-time inference, over 27 FPS, on the Ambarella CV25A chip of 1.6GHz ARM Cortex-A53 CPU with computer vision engine. 

\begin{figure}[t]
    \centering
    \includegraphics[width=0.5\textwidth]{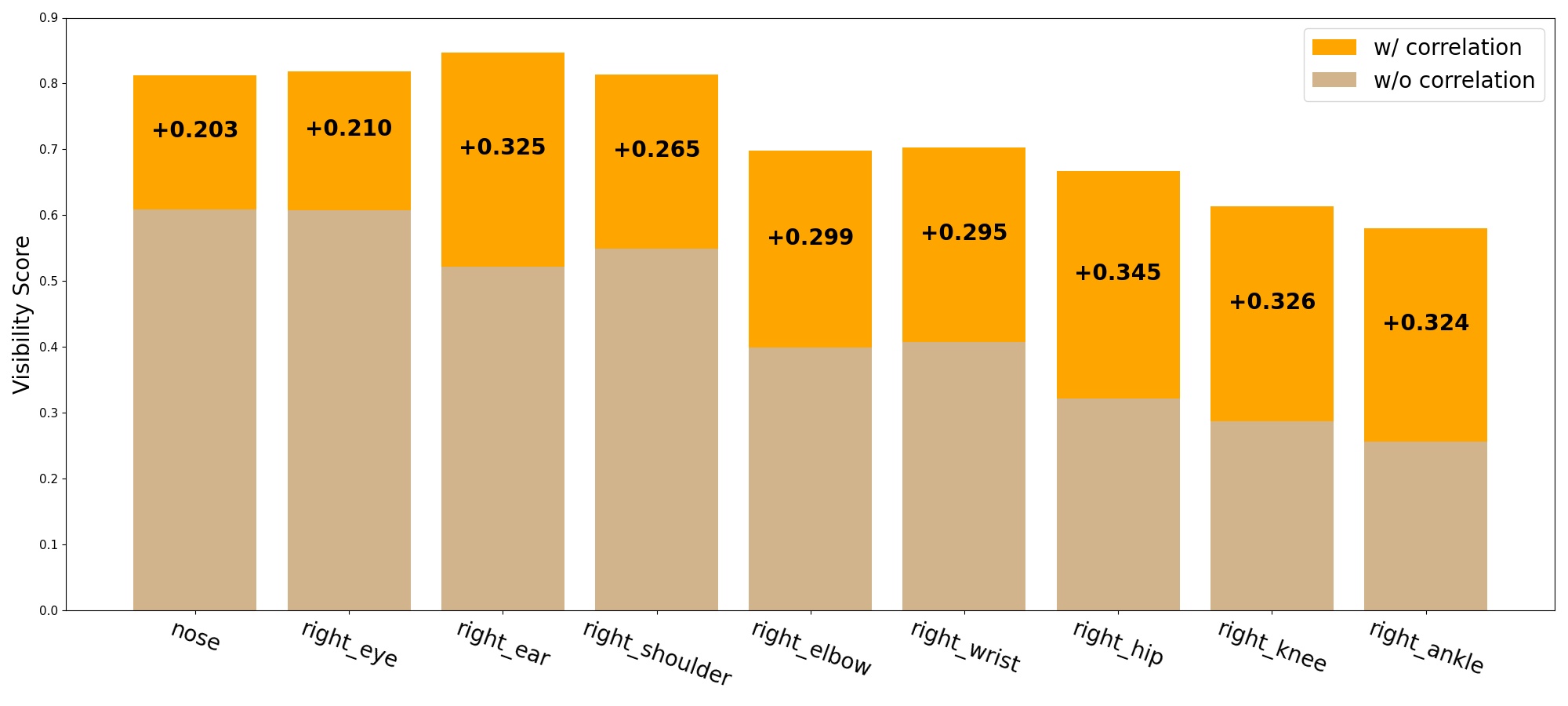}
    \caption{Ablation study of self-correlation loss on visibility scores.}
    \label{vis_score}
\end{figure}

\begin{figure*}
    \centering
    \includegraphics[width=1.0\textwidth]{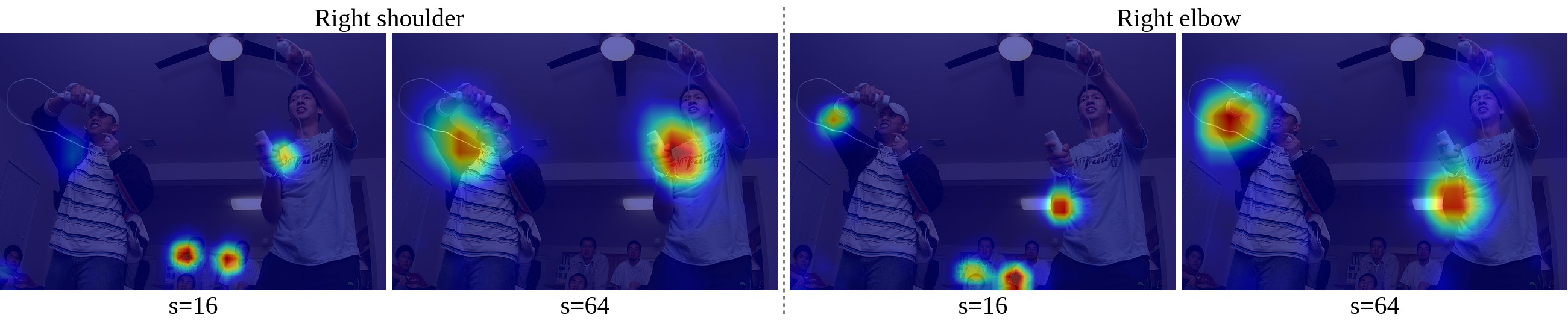}
    \caption{Visualization of visibility maps with different scales. A lower scale factor $s$ indicate higher resolution in visibility maps.}
    \label{vis_scale}
\end{figure*}

\begin{figure*}
    \centering
    \includegraphics[width=1.0\textwidth]{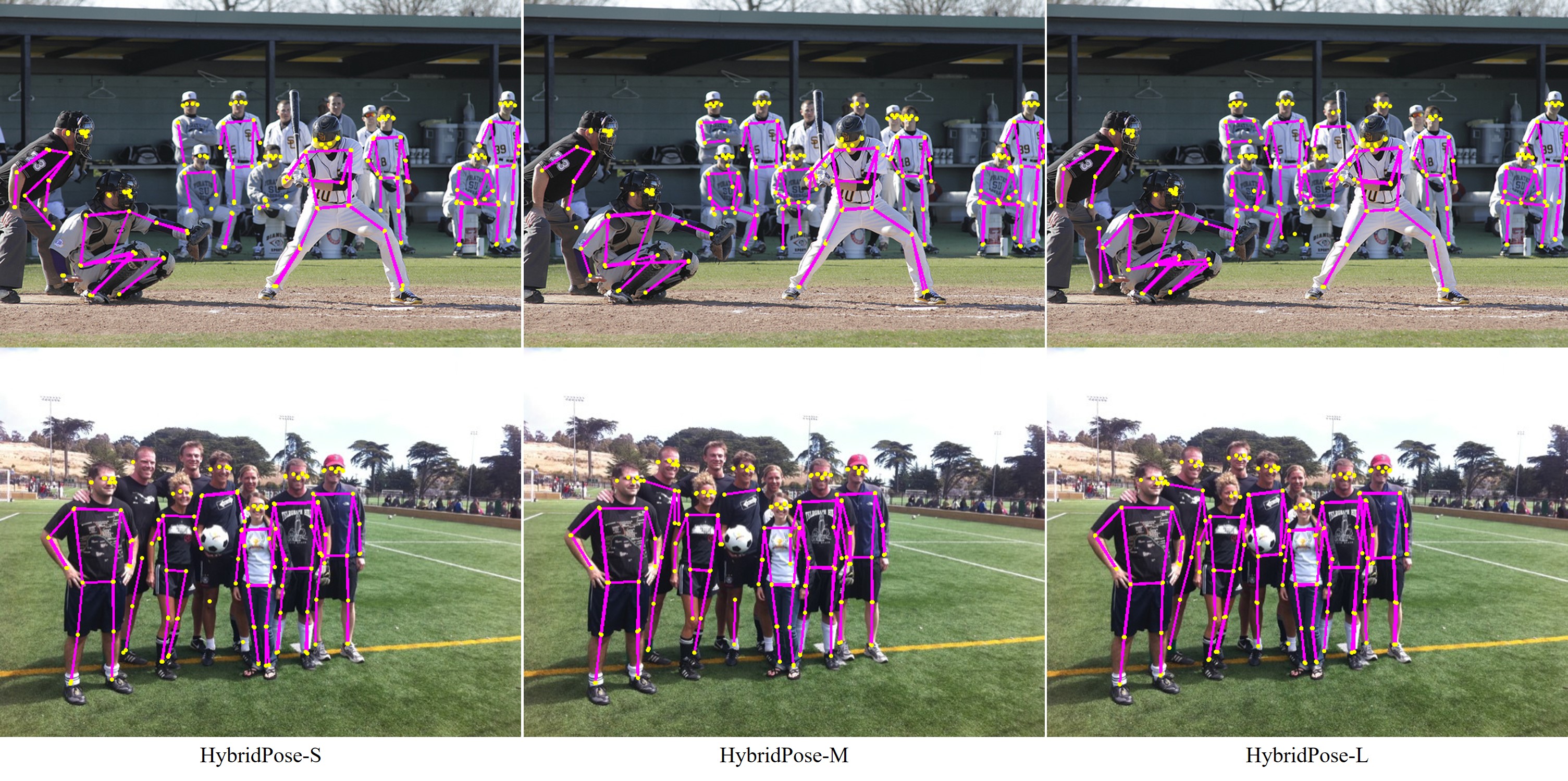}
    \caption{Comparisons of pose estimation and invisible keypoint filtering with HybridPose-S/M/L. To filter out invisible keypoints in scenes, visibility scores are utilized adopting Eq. \ref{visscore} with pose information.}
    \label{vis_size}
\end{figure*}

\subsection{Ablation studies}
\textbf{Self-Correlation Loss.} To validate the effectiveness of the self-correlation, we visualize visibility maps with keypoint coordinates according to adopting the self-correlation loss. As shown in left figures of Fig. \ref{corr_comp}, keypoint coordinates are uncorrelated with their corresponding visibility maps. It indicates that visible keypoints can be eliminated since their coordinates are positioned on out of visibility distributions. With the help of the self-correlation loss, the highest activations of visibility distributions are matched with their keypoint coordinates as shown in right figures of Fig. \ref{corr_comp}. Self-correlation loss quantitatively shows a similar tendency. To validate it, we measure the distance in the featuremap space between peak values of visibility distributions and their regression results. Self-correlation loss reduces the average distance across all keypoint labels between them from 5.7 to 4.9. To be specific, the distance is drastically reduced in the lower part of the body as shown in Fig. \ref{vis_dist}. It indicates that the highest activations of visibility distributions are center-oriented with their corresponding regression results. In addition, we measure visibility scores of predicted keypoint coordinates according to self-correlation loss. As described in Fig. \ref{vis_score}, visibility scores of each visible keypoint are inclined. Overall, average visibility scores of all keypoint labels increase from 0.4332 to 0.7289. Thus, HybridPose is capable of estimating valid visibility scores of each keypoint. Moreover, visibility values are narrowly distributed while being clustered in corresponding keypoint coordinates.

\textbf{Visibility Map.} HybridPose yields visibility maps in multiple resolutions. It indicates that each scale of the visibility maps contains different sizes of objects. As shown in Fig. \ref{vis_scale}, visibility distributions of small-sized objects appear in relatively high-resolution visibility maps and vice versa. Thereby, multi-scale predictions alleviate overlaps of visibility distributions. In addition, the model size affects visibility predictions similar to the pose estimation performance. Fig. \ref{vis_size} shows qualitative comparisons of pose estimation and invisible keypoint elimination against the model size of HybridPose. HybridPose-L is capable of detecting occluded person and filtering out invisible keypoints. Moreover, HybridPose-S shows the competitive performance compared with larger models while the performance degradation is discovered in terms of invisible keypoint elimination.

\section{Conclusion}
\label{sec:conclusion}
In this paper, we propose a HybridPose for the MPPE to mutually overcome each drawback of both regression-based and heatmap-based approaches by maximizing their strengths. To be specific, HybridPose simultaneously yields human bounding boxes, keypoint coordinates, and visibility maps in a single step. The human bounding boxes and keypoint coordinates are identified along with each person by sharing the same grid cell. Furthermore, the visibility maps facilitate HybridPose to handle invisible keypoints in a scene. In addition, we introduce the self-correlation loss to inject spatial dependencies between the keypoint coordinates and visibility maps since these are estimated in a independent phase. Consequently, HybridPose achieves a single-stage multi-person pose estimation while handling invisible keypoints. Moreover, we provide extensive experiments on the driver monitoring system and the edge device to demonstrate generalizability.

{\small
\bibliographystyle{ieee_fullname}
\bibliography{egbib}
}

\end{document}